\documentclass{article}
\usepackage[preprint]{neurips_2026}

\usepackage{iftex}
\ifPDFTeX
  \usepackage[utf8]{inputenc}
\fi
\usepackage[T1]{fontenc}
\usepackage[pdfversion=1.7]{hyperref}
\usepackage{booktabs}
\usepackage{amsmath}
\usepackage{amssymb}
\usepackage{graphicx}
\usepackage{xcolor}
\usepackage{flafter}
\usepackage{placeins}

\definecolor{effortmax}{HTML}{4E5D89}
\definecolor{effortxhigh}{HTML}{337F7A}
\newcommand{\reasoningtag}[2]{\,{\footnotesize\textcolor{#1}{[#2]}}}

\title{AnalogLayoutWorld: An Open Benchmark\\
for Agentic Analog Layout Design}
\author{%
  \raisebox{-0.25\height}{\includegraphics[height=22pt]{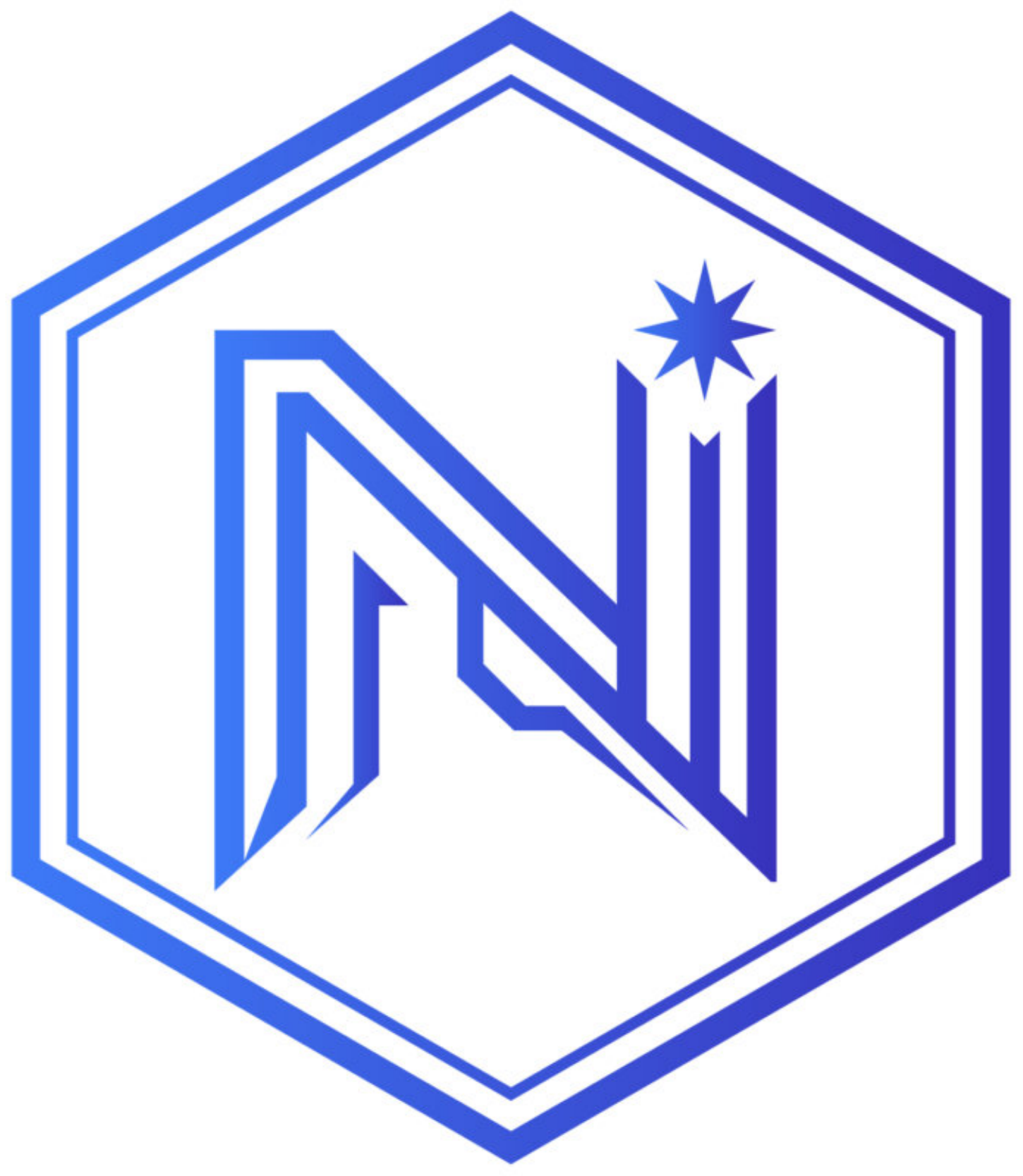}}%
  \hspace{0.6em}{\large\bfseries Nova-Silicon Team}\\[5pt]
  {\normalfont\small\hypersetup{hidelinks}%
   \href{mailto:xinxingyuan-sh@nova-silicon.com}{\texttt{xinxingyuan-sh@nova-silicon.com}}}%
}
\hypersetup{pdfauthor={Nova-Silicon Team}}

\begin{document}
\raggedbottom
\maketitle

\begin{abstract}
Coding agents have demonstrated strong capabilities in software development,
yet their ability to tackle long-horizon engineering tasks governed by
physical constraints remains less understood. We study this question through
analog circuit layout, which demands domain knowledge, geometric reasoning,
and iterative interaction with design and verification tools. We introduce
\textbf{AnalogLayoutWorld}, an open-source agent benchmark with 40
expert-reviewed tasks for end-to-end layout generation from fixed
transistor-level netlists. It combines an integrated design environment with
independent physical and post-layout electrical verification, as well as
objective layout quality metrics co-designed with analog layout experts.
Across 13 model--harness configurations, 91.1\% of attempts produce GDS
layouts, but only 21.3\% satisfy the benchmark's acceptance criteria. We also
evaluate Novara, an internally developed agent system, which achieves the
highest reported aggregate score. Our analysis links performance
differences to test-time scaling under fixed budgets, suggesting that
reasoning efficiency and effective use of verification feedback help sustain
design quality as task complexity grows.\footnote{Our benchmark repository will be open-sourced soon.}
\end{abstract}

\begingroup
\setlength{\intextsep}{3pt}
\vspace{12pt}
\begin{figure}[!ht]
  \centering
  \includegraphics[width=\linewidth]{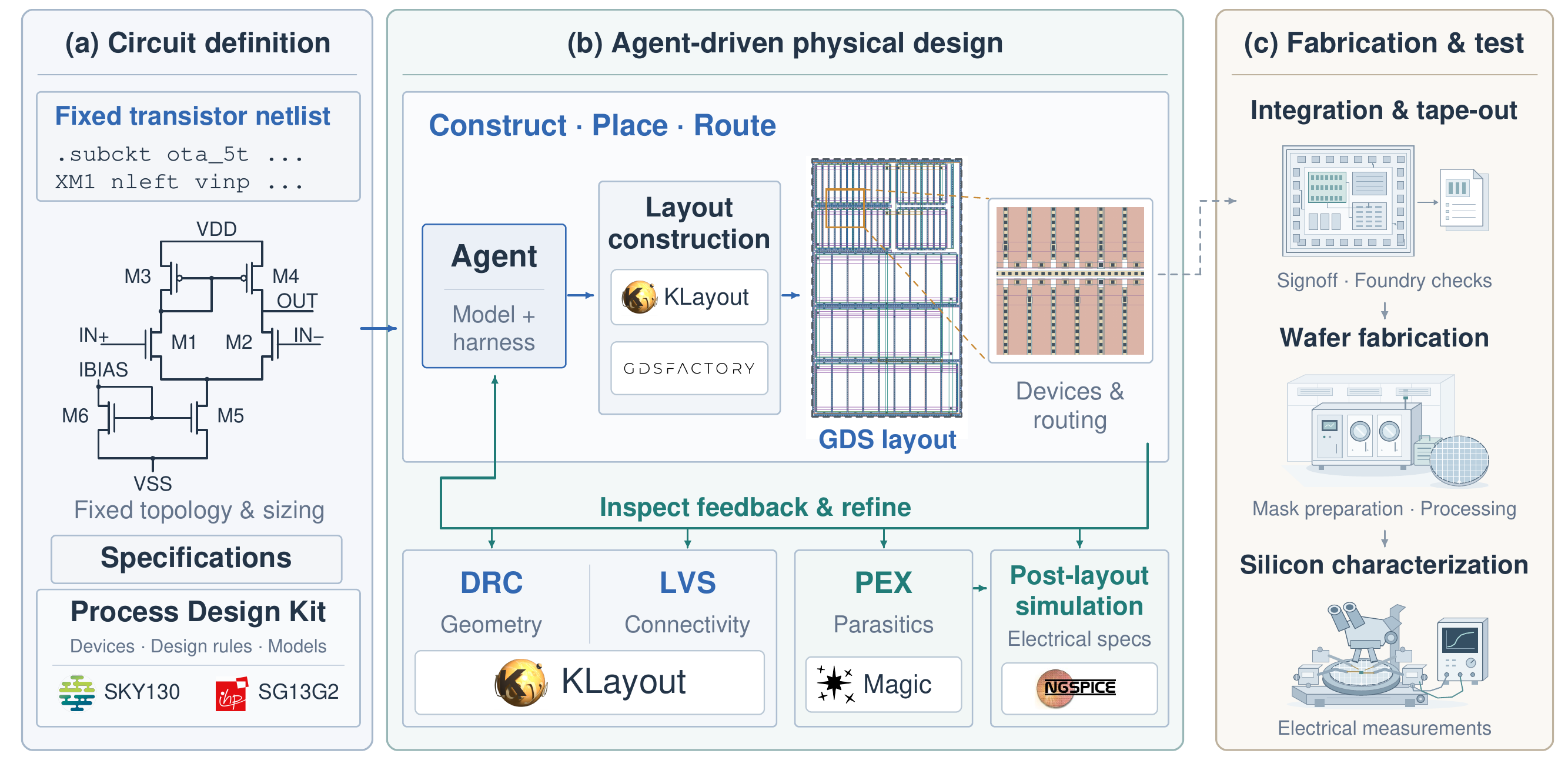}
  \caption{\textbf{Agentic analog layout design process.}
  (a) A fixed netlist, specifications, and process design kit define the task.
  (b) The agent constructs and refines layouts using layout tools and verification feedback.
  (c) Integration, fabrication, and silicon characterization provide physical validation.}
  \label{fig:task-demo}
\end{figure}

\clearpage
\endgroup

\section{Introduction}

\begingroup
\setlength{\textfloatsep}{10pt plus 2pt minus 2pt}
\begin{figure}[!b]
  \centering
  \setlength{\abovecaptionskip}{4pt}
  \includegraphics[width=\linewidth,trim=0bp 10bp 0bp 2bp,clip]{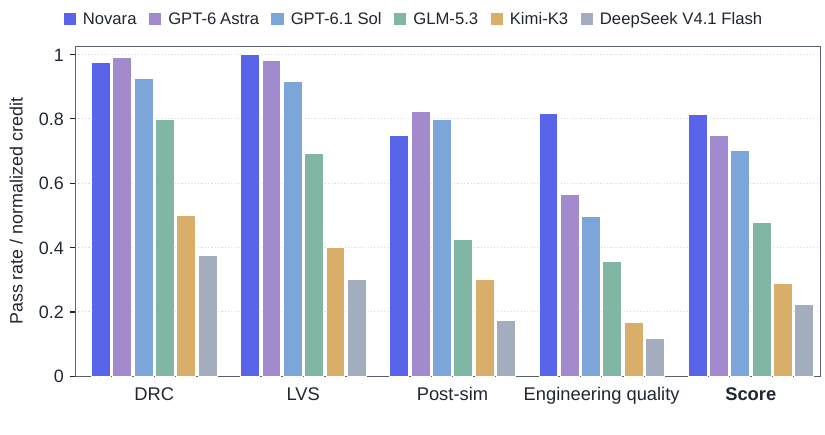}
  \caption{\textbf{Performance comparison of some agent systems across multiple verification metrics.}
  DRC, LVS, and Post-simulation report pass rates. 
  Score is the mean total over all 40 tasks.}
  \label{fig:main-comparison}
\end{figure}

General coding agents increasingly tackle software and terminal
tasks~\citep{jimenez2024swebench,merrill2026terminalbench}.
Recent foundation models advance coding, tool use, and scientific
reasoning~\citep{openai2026gpt6astra,anthropic2026claudeopus55,deepseekai2026deepseekv4,interns1team2025interns1},
motivating evaluation of agent reliability in long-horizon engineering
under physical constraints.

Analog layout requires domain knowledge and geometric reasoning to turn
fixed netlists into device placements and interconnects through iterative
verification (Figure~\ref{fig:task-demo}). Even layouts that pass physical
checks and post-layout simulation require separate assessment of compactness,
latch-up susceptibility, and electromigration
reliability~\citep{johnson2015latchup,jerke2014mission}.

Recently, many LLM-based methods try to address analog circuit design~\citep{hammoud2024glayout,li2026tlgrpo,zhang2026panda,liu2024layoutcopilot}.
However, limited task coverage and proprietary dependencies in some systems
motivate broader, objective evaluation of general-purpose agents.

We introduce \textbf{AnalogLayoutWorld}, an open benchmark of 40
expert-reviewed fixed-netlist layout tasks for agentic analog layout design. Our environment provides preinstalled open-source EDA
software and concise documentation, without benchmark-specific
typed tools or customized MCP interfaces. Agents can freely determine their own design
and verification workflows using the native execution tools provided by their harnesses. Independent grading assesses physical and
post-layout electrical correctness, compactness, and engineering quality
using objective metrics co-designed with engineers.

Across 13 model--harness configurations, 91.1\% of attempts produce GDS, but
only 21.3\% meet all acceptance criteria. Our in-house developed agent system Novara ranks
highest overall (Figure~\ref{fig:main-comparison}). Correlation analysis
reveals opposing patterns: circuit scale correlates strongly with GPT-series
resource use but less strongly with scores, whereas most analyzed open-source agents show the reverse.
Our findings suggest that most agents can still benefit from test-time
scaling. Under fixed budgets, however, realized gains depend on reasoning
efficiency and the ability to effectively convert environmental feedback
into verified improvements.

Finally, we summarize our key contributions as follows:
\begingroup
\setlength{\leftmargini}{10pt}
\setlength{\topsep}{2pt}
\setlength{\partopsep}{0pt}
\setlength{\itemsep}{5pt}
\setlength{\parsep}{0pt}
\begin{itemize}
  \item We introduce AnalogLayoutWorld, an open-source benchmark for agentic
  analog layout design with 40 expert-reviewed tasks, which combines an
  end-to-end open-source EDA software stack with independent physical, electrical, and
  layout-quality evaluation.
  \item We empirically characterize how iterative verification and test-time
  scaling relate to agent performance, highlighting the importance of
  reasoning efficiency and the ability to effectively convert environmental
  feedback into verified improvements within the available budget.
\end{itemize}
\endgroup

\FloatBarrier
\endgroup
\section{Related Work}

\subsection{Automated Analog Layout Design}

Analog layout automation seeks to translate circuit netlists and design
constraints into layouts that meet physical and electrical requirements.
MAGICAL~\citep{xu2019magical} and ALIGN~\citep{dhar2021align} automate
device generation and constraint-driven placement and routing.
LLM-based approaches extend this automation to language-to-layout generation
with GLayout~\citep{hammoud2024glayout} and interactive refinement with
LayoutCopilot~\citep{liu2024layoutcopilot}.
ICPI~\citep{liu2026icpi} incorporates simulation feedback into refinement,
while LAMP~\citep{xu2026lamp} targets adaptive placement.
Beyond individual layouts, \citet{chu2026codebehindlayout} study reusable
generator development, and PANDA~\citep{zhang2026panda} and
OPERA~\citep{saxena2026opera} integrate layout into broader design flows.
ViraStack~\citep{cadence2026virastack} brings agent orchestration into
commercial EDA, while \citet{yoshioka2026vibetapeout} examines agent-driven
design through post-layout verification. Despite this progress, evaluations remain
concentrated on selected circuit families and specialized workflows, with
proprietary dependencies in some systems, leaving the broader reliability and
reproducibility of autonomous layout design insufficiently established.

\subsection{Agent and IC Design Benchmarks}

Recent agentic benchmarks~\citep{liu2024agentbench,jimenez2024swebench,zhou2024webarena,xie2024osworld,merrill2026terminalbench,marten2026terminalbench4}
assess coding, tool use, and long-horizon task
completion across diverse domains and
environments.
Within IC design, digital benchmarks range from functionally verified RTL
generation~\citep{liu2023verilogeval} to workload-to-GDS design~\citep{chipagents2026buildchip}.
Analog circuit datasets and benchmarks address transistor-level design:
Analog Design Bench~\citep{analogdesignbench2026} evaluates netlist
design and simulation-based specification closure, while
LOADBench~\citep{leibl2026loadbench} provides circuit data and evaluation
tasks for topology and device sizing.
TapeoutBench~\citep{tapeoutbench2026} reports mixed chip-design activities,
including layout repair and routing.
Our work evaluates LLM agents on end-to-end analog layout
generation through an open EDA environment and independent verification,
assessing their capabilities in long-horizon, domain-specific problem solving
and iterative optimization using environmental feedback.

\section{AnalogLayoutWorld}
\label{sec:benchmark}

\begin{figure}[t]
  \centering
  \includegraphics[width=\linewidth]{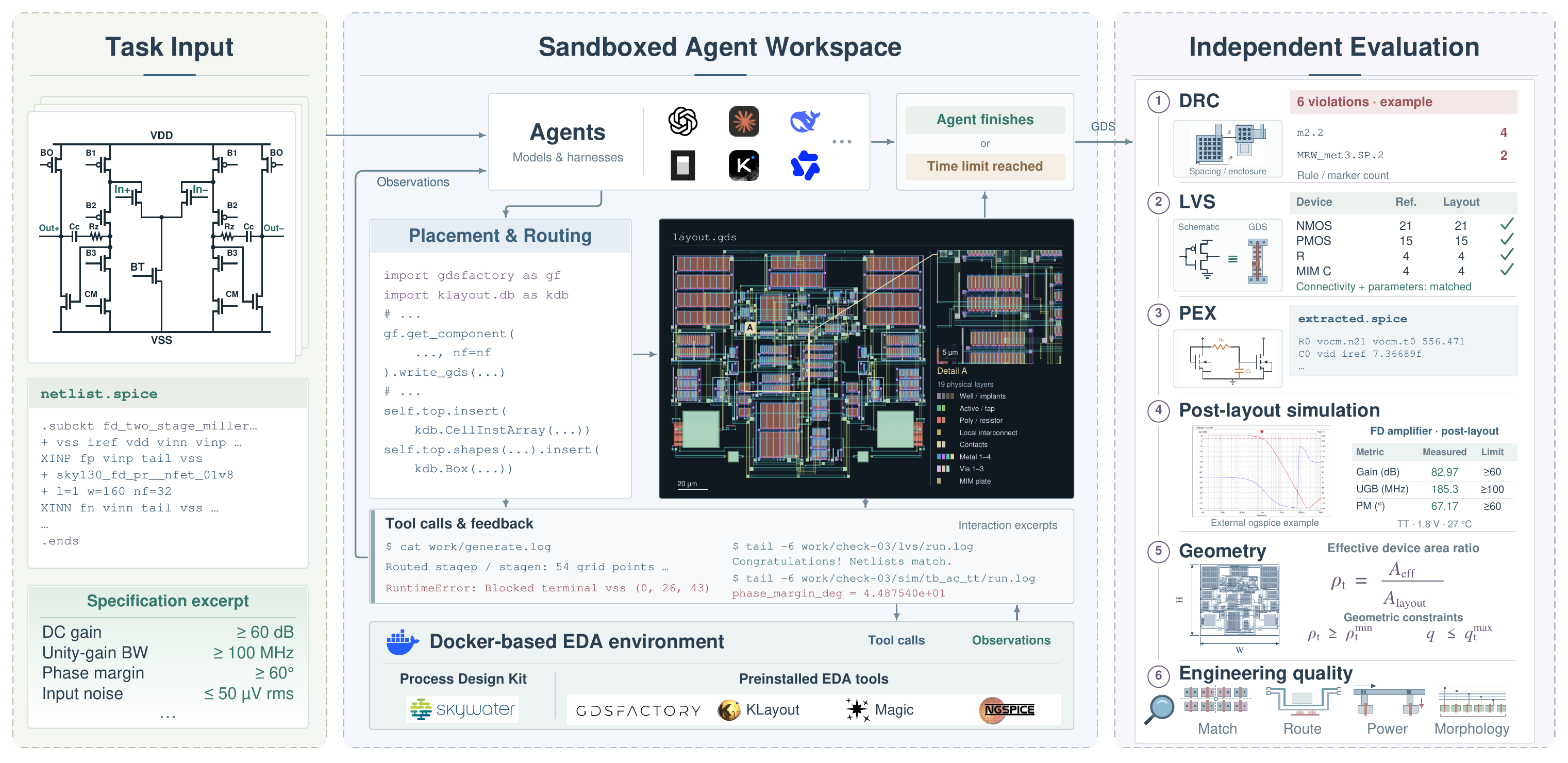}
  \caption{\textbf{Overview of AnalogLayoutWorld.}
  Agents receive fixed circuit inputs and interact with a preconfigured EDA
  workspace through the native tools from harnesses. After
  termination or timeout, the
  submitted GDS undergoes independent physical, electrical, and layout-quality
  assessment. The waveform is cropped from KiCad~\citep{kicadschematic9}.}
  \label{fig:overview}
\end{figure}

\subsection{Task Collection}
We construct AnalogLayoutWorld, a suite of 40 analog layout tasks, drawing on
circuits from \citet{analogdesignbench2026} and \citet{leibl2026loadbench}, our own
designs, and other open-source projects. The tasks cover
amplifiers, references, power-management circuits, sampling circuits, and
oscillators. Each task combines a fixed transistor-level SKY130 netlist with electrical
specifications and an executable validation procedure.

Analog circuit experts contribute to task selection and to the design and review
of netlists, specifications, and validation procedures. All circuits pass pre-layout simulation, establishing a verified electrical baseline for subsequent
layout generation.

\subsection{Task Format and Agent Environment}
We provide a flexible EDA environment with preinstalled open-source software and concise
interface documentation. Agents write their own scripts and choose how to
construct, verify, and revise their layouts using tool feedback.
Each task provides a fixed transistor-level SPICE netlist, electrical
specifications, implementation constraints, and public simulation testbenches.
Task instructions and documentation are available in both English and Chinese.

Built on Harbor~\citep{harbor}, our benchmark uses a custom
Docker~\citep{dockerdocs} image with the SKY130 PDK~\citep{sky130docs}.
The toolchain includes gdsfactory~\citep{gdsfactorydocs} for layout generation,
KLayout~\citep{klayoutdocs} for layout editing and physical verification including DRC and LVS,
Magic~\citep{magicdocs} for parasitic extraction (PEX), and
ngspice~\citep{ngspicedocs} for circuit simulation. Specifically, design rule checking (DRC) checks geometric constraints such as minimum wire
widths and spacings, while layout-versus-schematic (LVS) checking verifies
that devices and connections extracted from the layout match the reference
netlist.

Each attempt has a three-hour wall-clock budget for layout generation,
debugging, and self-checks. Agents must save the current layout to
\texttt{submission/layout.gds}, with the prescribed top cell and ports, and
retain generation code under \texttt{submission/src/}. Official grading begins
after agent termination or timeout, outside this execution budget.

Note that we provide no shared benchmark-specific typed tools or Model Context Protocol
(MCP) interfaces. Agents access the preinstalled software and usage documentation
through the native execution tools provided by their harnesses, deciding which
checks to run and how to revise their designs. The evaluation containers run
in a network-isolated environment. Together, these settings support open-ended
exploration in an isolated sandbox, subject to the task constraints.

\subsection{Evaluation}
\label{sec:evaluation}
After termination, we independently evaluate the submitted GDS in a fresh
directory within the same task container. Trusted references, testbenches,
and grading code are introduced separately; agent-reported results do not
determine the score. Evaluation comprises DRC, LVS, PEX and post-layout
simulation, together with geometric and engineering-quality measurements.

\paragraph{Geometry and eligibility.}
Compact layouts reduce silicon cost and can shorten interconnects, limiting
parasitic resistance and capacitance. We measure compactness with the
\emph{effective device-area ratio}
\begin{equation}
\rho_t=\frac{A_{\mathrm{eff}}}{A_{\mathrm{layout}}},
\qquad A_{\mathrm{layout}}=WH,
\qquad q=\frac{\max(W,H)}{\min(W,H)}.
\label{eq:device-area}
\end{equation}
Here $W,H$ bound the layout's manufacturing geometry, and $A_{\mathrm{eff}}$
sums the areas of LVS-recognized device regions corresponding to the reference circuit.
The geometric indicator $I_{\mathrm{GEO}}$ requires valid measurements,
$q\le q_t^{\max}$, and $\rho_t\ge\rho_t^{\min}$; the aspect bound discourages
elongated blocks that are difficult to integrate with neighboring circuits.
The morphology indicator $I_{\mathrm M}$ denotes passing a screen for
degenerate bus-and-row layouts. Let $I_{\mathrm G}=1$ if the agent submits a
GDS file at the required path and $I_{\mathrm L}=1$ if LVS passes; each is
zero otherwise. Eligibility is then
\begin{equation}
E_t=I_{\mathrm G}I_{\mathrm L}I_{\mathrm{GEO}}I_{\mathrm M}.
\label{eq:layout-eligibility}
\end{equation}
Required geometry and morphology evidence must be available for $E_t=1$.

\paragraph{Scoring.}
Let $I_{\mathrm D}$ indicate passing DRC. For layout $\mathcal L$ on task $t$,
the score is
\begingroup
\thinmuskip=2mu\medmuskip=2mu\thickmuskip=3mu
\makeatletter
\@fleqntrue\@mathmargin=0pt
\renewenvironment{equation}{%
  \incr@eqnum\mathdisplay@push
  \st@rredfalse\global\@eqnswtrue\mathdisplay{equation}%
}{%
  \endmathdisplay{equation}\mathdisplay@pop\ignorespacesafterend
}
\makeatother
\begin{equation}
S_t(\mathcal L)=\underbrace{\omega_{\mathrm G}I_{\mathrm G}
 +I_{\mathrm L}\bigl(\omega_{\mathrm L}+\omega_{\mathrm D}I_{\mathrm D}\bigr)}_{\text{Submission and physical verification}}
 +E_t\Biggl[\underbrace{\omega_{\mathrm A}\min\!\left\{1,\frac{\rho_t}{\rho_t^\star}\right\}^{\gamma}}_{\text{Compactness}}
 +\underbrace{\omega_{\mathrm Q}Q_t}_{\text{Engineering quality}}
 +\underbrace{\omega_{\mathrm S}F_t}_{\text{Electrical verification}}\Biggr].
\label{eq:layout-score}
\end{equation}
\endgroup
Here $\rho_t^\star$ sets the full-credit area-ratio target and $\gamma>0$ controls
compactness sensitivity. $Q_t\in[0,1]$ aggregates device matching,
critical-net routing, and power-delivery quality, including a coupling
penalty. $F_t\in[0,1]$ is the verified passing fraction of a fixed electrical
verification manifest, and is zero without successful PEX. DRC credit requires
passing LVS; the three terms gated by $E_t$ do not additionally require DRC
or all-case simulation success. Complete acceptance additionally requires
passing DRC and all electrical cases, with complete evidence; partial scores preserve
verified progress.
Nonnegative weights with $\sum_k\omega_k=1$ ensure $S_t\in[0,1]$; task scores
are averaged equally. For detailed metric definitions and engineer-specified parameter values, see Appendix~\ref{app:evaluation}.

\FloatBarrier

\section{Main Results}
\label{sec:main-results}

\subsection{Experimental Setup}
We evaluate 13 model--harness configurations, with reasoning settings listed
in Table~\ref{tab:main-results}. The GPT models~\citep{openai2026gpt61sol,openai2026gpt6astra} use
Codex~\citep{codexcli}; GLM models~\citep{glm5team2026glm5vibecodingagentic} use OpenCode~\citep{opencode}, with
GLM-5.3 and its Flash variant also evaluated in our InternBootcamp
harness~\citep{li2026internbootcamp,li2026internagentharness}.
DeepSeek models~\citep{deepseekai2026deepseekv4,deepseekai2026deepseekv41flash} use Claude Code~\citep{claudecode}, and DeepSeek V4.1 Flash
is additionally evaluated with DeepSeek Harness (DSH)~\citep{deepseek-harness2026}.
Kimi models~\citep{moonshot2026kimik26,kimiteam2026kimik3} use
Kimi Code CLI~\citep{kimicode}, and Qwen3.8 Flash Next~\citep{qwen2026design}
uses Qwen Code~\citep{qwencode}.
Except for the GPT models accessed through an API, all models are deployed
on our own cluster using SGLang~\citep{zheng2024sglang}.
Using the environment, three-hour agent budget, and independent evaluation
protocol described in Section~\ref{sec:benchmark}, we run three independent
trials per task. For the figures and case studies, we use OpenCode for GLM models and DSH for DeepSeek V4.1 Flash; Table~\ref{tab:main-results} includes all evaluated configurations.

\paragraph{Novara Agent System.}
We also evaluate Novara, an in-house automated agent system specifically designed for analog circuit
layout and developed by our Nova-Silicon team~\citep{li2026agentorchestration}.
It supports multiple process technologies, is deployed entirely on our own
cluster, and uses no external APIs.

\subsection{Overall Results}
Table~\ref{tab:main-results} shows substantial variation in layout-generation
performance across the evaluated systems. Novara achieves the highest aggregate
score (0.8154), followed by GPT-6 Astra (0.7483).
The GPT configurations attain substantially higher physical and electrical
verification rates than the other general-purpose agents, while
several configurations struggle to produce circuit-equivalent layouts.
For GPT agents, violations of geometric and morphological constraints, including
aspect-ratio limits, account for most of the gap between simulation pass rates
and complete acceptance.
Producing a layout artifact alone therefore does not establish successful
completion of the task.

Results in Table~\ref{tab:main-results} also indicate that the relative strengths of agent systems depend on which aspect of layout quality is
examined. GPT-6 Astra leads in DRC and all-case post-layout simulation,
whereas Novara leads in LVS and credited engineering quality.
Figure~\ref{fig:main-comparison} makes this distinction visible: the aggregate
ranking reflects both verified circuit behavior and engineering quality,
whose relative strengths differ across agent systems. The ordering of general-purpose agents is broadly consistent with
expectations based on their general coding capabilities.

\begin{table}[!htbp]
  \centering
  \caption{\textbf{Main results on AnalogLayoutWorld.}
  Results are averaged over three independent repetitions per task.
  DRC, LVS, and SIM are pass rates over all task attempts;
  SIM requires all prescribed post-layout cases to pass.
  Acc. reports the fraction of task attempts meeting all benchmark acceptance criteria.
  Area is the effective device-area ratio. Toks counts cumulative recorded input and output tokens.
  Tool calls count tool invocations. Brackets show reasoning settings or deployment.
  Bold marks column maxima for outcomes and minima for resource use.}
  \label{tab:main-results}
  \fontsize{8.5}{10}\selectfont
  \setlength{\tabcolsep}{1.3pt}
  \renewcommand{\arraystretch}{1.14}
  \begin{tabular*}{\linewidth}{@{\extracolsep{\fill}}llrrrrrrrrr@{}}
    \toprule
    Model / System & Harness & \shortstack{DRC\\(\%)} & \shortstack{LVS\\(\%)} & \shortstack{SIM\\(\%)} & \shortstack{Acc.\\(\%)} & \shortstack{Area\\(\%)} & \shortstack{Quality\\$E_tQ_t$} & \shortstack{Tool\\Calls $\downarrow$} & \shortstack{Toks \\ (M)$\downarrow$} & \shortstack{\textbf{Avg.}\\\textbf{Score}} \\
    \midrule
    \multicolumn{2}{l}{Novara Agent System\reasoningtag{effortmax}{locally deployed}} & 97.5 & \textbf{100.0} & 75.0 & \textbf{72.5} & 52.2 & \textbf{0.816} & \textemdash & \textemdash & \textbf{0.8154} \\
    DeepSeek V4 Pro\reasoningtag{effortmax}{max} & Claude Code & 23.3 & 23.3 & 12.5 & 3.3 & 11.3 & 0.049 & 226.9 & 23.8 & 0.1504 \\
    DeepSeek V4.1 Flash\reasoningtag{effortmax}{max} & Claude Code & 41.7 & 29.2 & 17.5 & 8.3 & 35.0 & 0.086 & 134.0 & 13.2 & 0.1974 \\
    DeepSeek V4.1 Flash\reasoningtag{effortmax}{max} & DSH & 37.5 & 30.0 & 17.5 & 9.2 & 44.5 & 0.118 & 133.1 & 14.7 & 0.2246 \\
    GLM-5.2 FP8\reasoningtag{effortmax}{max} & OpenCode & 10.0 & 10.0 & 5.0 & 0.0 & 21.3 & 0.000 & 242.3 & 18.3 & 0.1075 \\
    GLM-5.3\reasoningtag{effortmax}{max} & OpenCode & 80.0 & 69.2 & 42.5 & 35.0 & 35.6 & 0.358 & 264.8 & 34.5 & 0.4779 \\
    GLM-5.3\reasoningtag{effortmax}{max} & InternBootcamp & 55.0 & 65.0 & 45.0 & 27.5 & 49.8 & 0.350 & 253.3 & 30.3 & 0.4632 \\
    GLM-5.3 Flash\reasoningtag{effortmax}{max} & OpenCode & 44.2 & 33.3 & 22.5 & 14.2 & 27.8 & 0.146 & 239.1 & 31.7 & 0.2555 \\
    GLM-5.3 Flash\reasoningtag{effortmax}{max} & InternBootcamp & 40.0 & 30.0 & 22.5 & 12.5 & 26.3 & 0.117 & 310.8 & 38.0 & 0.2378 \\
    Kimi-K2.6\reasoningtag{effortmax}{max} & Kimi Code & 0.8 & 0.0 & 0.0 & 0.0 & \textemdash & 0.000 & 235.0 & 23.0 & 0.0958 \\
    Kimi-K3\reasoningtag{effortmax}{max} & Kimi Code & 50.0 & 40.0 & 30.0 & 20.0 & 37.5 & 0.167 & 142.0 & 12.2 & 0.2900 \\
    Qwen3.8 Flash Next\reasoningtag{effortxhigh}{xhigh} & Qwen Code & 40.0 & 41.7 & 25.0 & 14.2 & 32.9 & 0.140 & 217.4 & 21.6 & 0.2633 \\
    GPT-6 Astra\reasoningtag{effortmax}{max} & Codex & \textbf{99.2} & 98.3 & \textbf{82.5} & 70.8 & 59.5 & 0.566 & \textbf{111.8} & \textbf{11.2} & 0.7483 \\
    GPT-6.1 Sol\reasoningtag{effortmax}{max} & Codex & 92.5 & 91.7 & 80.0 & 62.5 & \textbf{60.5} & 0.498 & 131.6 & 14.5 & 0.7023 \\
    \bottomrule
  \end{tabular*}
\end{table}

\FloatBarrier
\section{Analysis}
\label{sec:analysis}
Beyond aggregate scores, we analyze verification progress, task-scale
sensitivity, and test-time resource use. By relating circuit scale to both
final quality and resource expenditure, we seek a quantitative explanation 
for the performance gap between the evaluated open-weight configurations
and GPT-based agents from the perspective of test-time scaling. The central question is whether systems can sustain
productive iteration as task demands grow under a fixed time budget.

\subsection{Progress through Verification Feedback}
\label{sec:feedback-refinement}
\begin{figure}[!htbp]
  \centering
  \includegraphics[width=\linewidth]{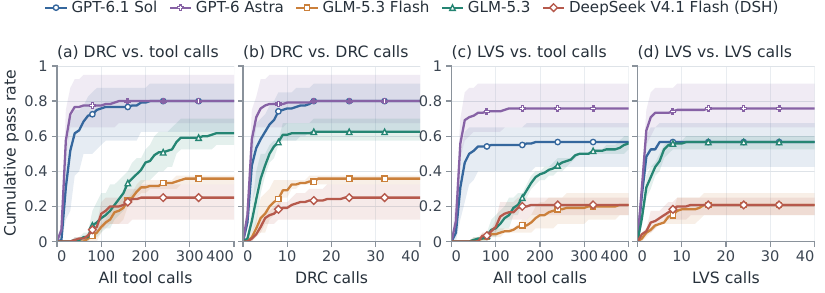}
  \caption{\textbf{Verification progress during agent interaction.}
  Curves show the fraction of tasks with a recovered passing self-check by a
  given budget, averaged over three independent runs; shading shows the range
  across runs. The left pair shows DRC and the right pair LVS, indexed by all
  tool calls and the respective verification calls. All panels share a
  vertical scale from 0 to 1.}
  \label{fig:verification-progress}
\end{figure}

We first examine recovered DRC and LVS self-checks to track how verification
progress accumulates during agent interaction,
as demonstrated in Figure~\ref{fig:verification-progress}.

\paragraph{Agents make progress through repeated verification.}
Agents often require repeated verification attempts to reach a passing
self-check. Among trajectories with a recovered passing self-check,
90.3\% of DRC first passes and 66.8\% of LVS first passes occur after the
first verification call. The gains beyond the initial check show that
continued interaction can produce additional verified progress. Repeated
verification is therefore a substantive part of trajectories with recovered
passing self-checks, supporting productive design--verification loops of iterative optimization.

\paragraph{Verification progress remains uneven.}
The amount of verified progress achieved differs substantially across
configurations. Figure~\ref{fig:verification-progress} indexes progress by
both total tool calls and verification calls, distinguishing overall
interaction effort from repeated checking. At common budgets, the systems
reach different fractions of tasks with passing checks, and many
trajectories still lack a recovered pass within the displayed range.
Thus, the opportunity to iterate does not yield the same progress for every
system, motivating a closer examination of task demands and resource use, which we discuss in the following sections.

\FloatBarrier
\subsection{Task Scale and Difficulty}
\label{sec:task-complexity}
\begin{figure}[!t]
  \centering
  \includegraphics[width=\linewidth]{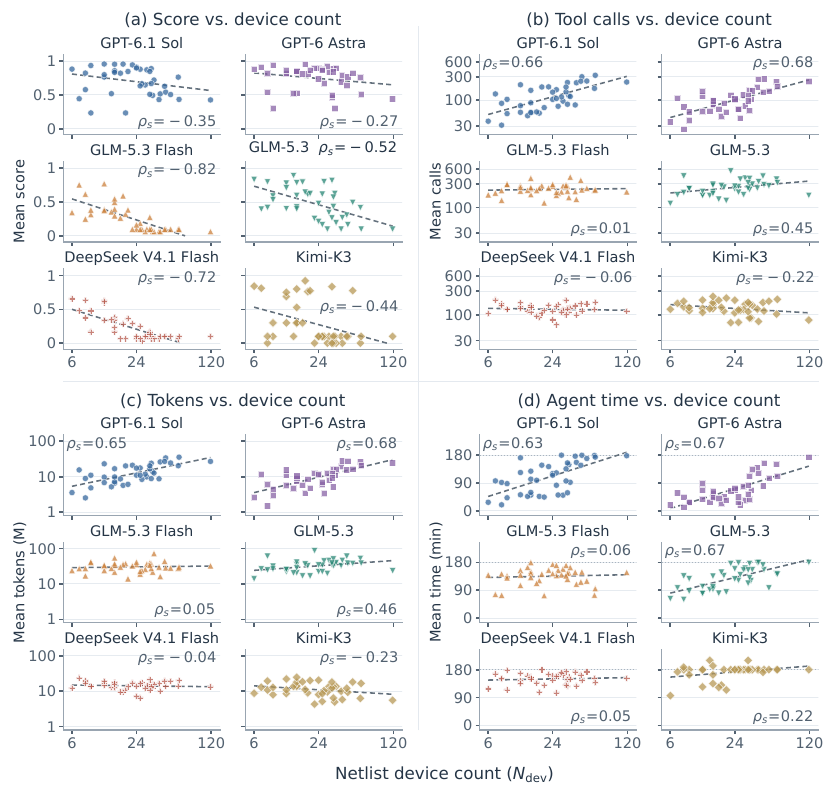}
  \caption{\textbf{Task scale, design quality, and resource use.}
  Each point is a task-level mean for one configuration. Groups show
  (a) score, (b) tool calls, (c) reported input-plus-output tokens in millions,
  and (d) agent time in minutes. Device count, calls, and tokens use logarithmic
  axes; in-panel $\rho_s$ denotes Spearman rank correlation. Dashed lines show
  least-squares trends in the displayed coordinates, restricted to $[0,1]$
  for scores.}
  \label{fig:task-scale}
\end{figure}

We next examine how the scale of the input circuit relates to final layout
quality. Figure~\ref{fig:task-scale}(a) compares source-netlist device count
with independently evaluated final scores.

\paragraph{Task scale reflects empirical difficulty.}
Device count provides an agent-independent measure of circuit scale:
additional devices expand the placement and connectivity requirements that
an agent must coordinate. Averaging final scores across the six plotted
configurations yields a negative task-level correlation with device count
($\rho_s=-0.73$), indicating that larger circuits tend to be more difficult
for the evaluated systems. See Appendix~\ref{app:task-scale} for the
complementary net-count analysis, which shows similar trends. Together,
these results support circuit scale as a useful descriptor of empirical
task difficulty in AnalogLayoutWorld.

\paragraph{Scale sensitivity differs across systems.}
Increasing circuit scale affects some systems much more strongly than
others. GLM-5.3 Flash, GLM-5.3, DeepSeek V4.1 Flash, and Kimi-K3 exhibit
stronger negative correlations ($\rho_s=-0.82$ to $-0.44$), whereas
GPT-6.1 Sol and GPT-6 Astra show weaker associations
($-0.35$ and $-0.27$). Because all configurations face the same task
collection and scoring criteria, this contrast raises an important question:
why do the GPT configurations preserve quality better as circuit scale
increases? We examine their use of tools, tokens, and time next to understand
how they respond to these growing demands.

\subsection{Test-Time Scaling under Budget Constraints}
\label{sec:resource-use}
To understand these differences in scale sensitivity, we relate circuit
scale to both final scores and resource use (Figure~\ref{fig:task-scale}). The results indicate that 
GPT-series agents show stronger positive resource correlations but weaker
negative score correlations, whereas most open-weight configurations exhibit the
reverse. We examine whether differences in test-time scaling help explain
this contrast.

\paragraph{GPT-series agents scale interaction with task demands.}
GPT-6.1 Sol and GPT-6 Astra devote more interaction to larger circuits.
Device count correlates positively with their tool calls
($\rho_s=0.66$ and $0.68$) and elapsed time ($0.63$ and $0.67$), alongside
similarly positive token trends. Their recorded workloads therefore expand
with circuit scale, while final scores show comparatively smaller declines.
Finishing smaller tasks with time to spare leaves room for longer
design--verification loops as task demands grow.
Combined with the verification progress in
Section~\ref{sec:feedback-refinement}, this pattern is consistent with
productive test-time scaling through continued environment interaction:
larger tasks receive more effort, with smaller score declines than observed
for the open-weight agents.

\paragraph{Budget saturation limits room for further iteration.}
High expenditure on smaller circuits leaves less room to extend the
verification process as task demands grow. GLM-5.3 Flash, DeepSeek V4.1 Flash,
and Kimi-K3 already incur substantial time costs on smaller circuits and
exhibit weaker time--size associations. GLM-5.3 Flash reaches the
agent time limit in 51 of 120 attempts, compared
with none for GPT-6 Astra. GLM-5.3 is an intermediate case, with a strong
time--size correlation ($\rho_s=0.67$) but more modest growth in calls and
tokens. These patterns suggest that limited budget headroom can constrain
additional verification and repair. As task demands increase, several
open-weight configurations show weaker growth in interaction
and stronger score degradation, consistent with a budget-limited response
to increasing difficulty.

\paragraph{Efficiency and realized test-time scaling.}
The contrasting correlations suggest that the performance gap partly
reflects differences in realized test-time scaling. The GPT agents combine
stronger growth in effort with weaker score degradation as circuit scale
increases; most open-weight configurations already incur high costs on
smaller tasks, limiting room to extend iteration on harder ones. GPT-6 Astra
also achieves a mean score of 0.7483 with 11.2M reported tokens per task,
compared with 0.4779 and 34.5M for GLM-5.3
(Table~\ref{tab:main-results}), consistent with more effective use of
test-time resources. Together with the observed verification progress,
these findings suggest that most of the agents examined here can still
benefit from test-time scaling. However, under practical budget constraints,
its realized benefits depend on reasoning efficiency and the ability to
effectively convert environmental feedback into verified improvements
within the available budget.

\FloatBarrier
\section{Case Studies}
\label{sec:case-studies}
In this section, we present representative cases to examine agent behavior
and illustrate how verification and engineering-quality assessment shape
the final scores.

\subsection{Residual Failures in Strong Agents}
\label{sec:case-outcomes}
\begin{figure}[!htbp]
  \centering
  \includegraphics[width=\linewidth]{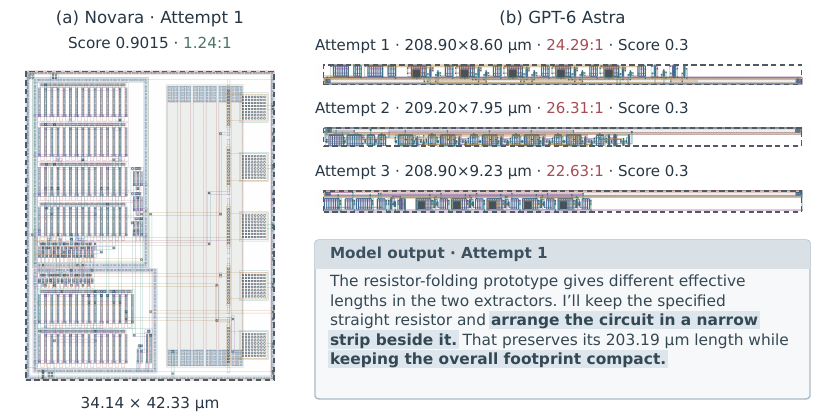}
  \caption{\textbf{Current-starved ring VCO submissions.}
  Novara's first attempt is compared with three independent GPT-6 Astra attempts. All four
  pass DRC, LVS, and both prescribed post-layout tests, but the Astra
  layouts exceed the aspect-ratio bound. Dashed outlines mark the bounding
  rectangles. The box quotes Astra's first attempt, highlighting its strip
  choice and compactness assessment.}
  \label{fig:case-vco}
\end{figure}

The current-starved ring VCO (Task~9) illustrates a physical-design limitation that
persists despite successful verification (Figure~\ref{fig:case-vco}).
All three GPT-6 Astra attempts pass DRC, LVS, and both prescribed
post-layout tests. Nevertheless, their aspect
ratios range from 22.63 to 26.31, exceeding the evaluation bound of 4.
Each receives 0.1 for submission, LVS, and DRC, totaling 0.3; geometric
ineligibility withholds the area, engineering-quality, and electrical
credits despite the passing simulations. In comparison, all Novara runs
pass the same checks; its first attempt achieves an aspect ratio of 1.24
and a score of 0.9015, demonstrating that
electrical acceptance and a compact block shape can be achieved together.

The recorded trajectories, including the model output in
Figure~\ref{fig:case-vco}, explain how this repeated outcome arose.
Each GPT-6 Astra attempt explores a folded resistor whose extracted length
fails LVS parameter matching, then adopts a straight resistor of
$203.19\,\mu\mathrm m$. This device dominates the block width even after
local placement improvements. In the second attempt, the agent also
constructs and verifies a clustered transistor placement before restoring
the earlier strip implementation. Thus, the failure is not an absence of
verification or geometric exploration: the agent makes local repairs but
does not find an implementation that jointly achieves electrical closure
and a suitable global shape. The three independent attempts expose a
remaining limitation even in a strong closed-source model.

\FloatBarrier
\subsection{Engineering Quality beyond Electrical Acceptance}
\label{sec:case-engineering}
\begin{figure}[!htbp]
  \centering
  \includegraphics[width=\linewidth]{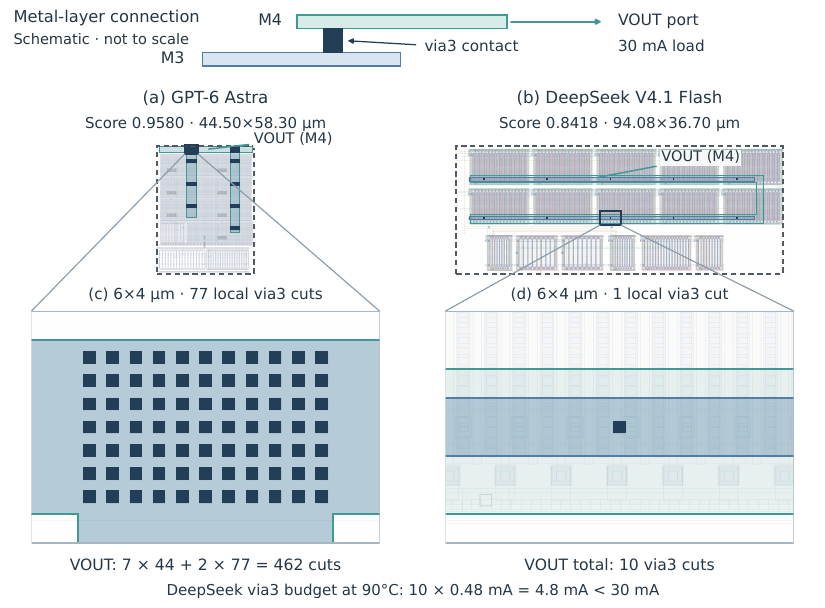}
  \caption{\textbf{Power-delivery geometry in the LDO.}
  The top schematic explains the M3--via3--M4 output connection.
  Colored overlays show the extracted VOUT geometry in the submitted GDS.
  Equal $6\times4\,\mu\mathrm m$ detail windows contain 77 and 1 local
  via3 cuts; the respective whole-net totals are 462 and 10. Complete layouts are fitted
  separately with their aspect ratios preserved.}
  \label{fig:case-ldo}
\end{figure}

The $30\,\mathrm{mA}$ PMOS-pass LDO (Task~17) shows why functional acceptance alone
does not establish interconnect adequacy (Figure~\ref{fig:case-ldo}).
The GPT-6 Astra and DeepSeek V4.1 Flash submissions both pass DRC, LVS,
and all prescribed post-layout simulations. Both expose VOUT on metal4,
but their output networks contain 462 and ten via3 cuts, respectively,
connecting metal3 to metal4. The DeepSeek layout has no metal5 bypass;
when the specified load is carried through its declared output access,
these ten cuts must collectively carry $30\,\mathrm{mA}$.
The public SkyWater technology
LEF~\citep{skywatertechlef} specifies a DC-average current budget of
$0.48\,\mathrm{mA}$ per via3 cut at a junction temperature of
$90^\circ\mathrm C$. Ten cuts therefore provide only $4.8\,\mathrm{mA}$
of aggregate capacity at this reference condition; at least 63 are necessary.

Table~\ref{tab:case-ldo-scores} separates the resulting score contributions.
Both layouts earn the full 0.7 submission, physical, and electrical
credits; their differences arise from compactness and engineering quality.
Within the latter, the normalized power-delivery subscore is 0.748 for
GPT-6 Astra and 0.199 for DeepSeek V4.1 Flash, with matching and routing
also contributing to the overall quality difference. The via-capacity
check supplies a physical basis for a quality distinction that the
prescribed functional tests do not expose. It is a necessary current-budget
screen at the stated reference condition, rather than complete
electromigration or lifetime sign-off. Additional compactness and circuit
correctness examples are demonstrated in Appendix~\ref{app:case-studies}.

\begin{table}[!htbp]
  \centering
  \caption{\textbf{LDO score decomposition.}
  Entries are weighted contributions to the total score. Both submissions
  pass DRC, LVS, and all prescribed electrical tests.}
  \label{tab:case-ldo-scores}
  \small
  \setlength{\tabcolsep}{3pt}
  \begin{tabular*}{\linewidth}{@{\extracolsep{\fill}}lrrrrrrr@{}}
    \toprule
    Submission & GDS & LVS & DRC & Area & Quality & SIM & Total \\
    \midrule
    GPT-6 Astra & 0.1000 & 0.1000 & 0.1000 & 0.1651 & 0.0929 & 0.4000 & 0.9580 \\
    DeepSeek V4.1 Flash & 0.1000 & 0.1000 & 0.1000 & 0.0956 & 0.0462 & 0.4000 & 0.8418 \\
    \bottomrule
  \end{tabular*}
\end{table}

\FloatBarrier
\section{Conclusion}
We presented AnalogLayoutWorld, an open-source benchmark for agentic analog
layout design with 40 expert-reviewed tasks, an end-to-end open-source EDA
software stack, and independent physical, electrical, and layout-quality
evaluation. Agents freely determine their design and verification workflows,
while assessment is grounded in submitted layouts. Our experiments reveal
a substantial gap between artifact generation and verified completion.
Analyses of verification progress, task scale, and resource use suggest that
most agents can benefit from test-time scaling, but gains under finite
budgets depend on reasoning efficiency and effective use of environmental
feedback. We discuss the scope and limitations of our work in Appendix~\ref{app:limitations}.
Future work will broaden circuit and process coverage and refine
layout-quality assessment through expert review and community feedback.

\section*{Acknowledgments}
We sincerely thank the authors of Analog Design Bench~\citep{analogdesignbench2026}
and LOADBench~\citep{leibl2026loadbench} for openly releasing the circuit data
used to construct part of our task suite. We also thank the developers of Harbor~\citep{harbor} and SGLang~\citep{zheng2024sglang} for
the evaluation and model-serving infrastructure, and the communities behind
KLayout~\citep{klayoutdocs}, gdsfactory~\citep{gdsfactorydocs},
Magic~\citep{magicdocs}, Netgen~\citep{netgendocs},
ngspice~\citep{ngspicedocs}, SKY130~\citep{sky130docs}, and
open\_pdks~\citep{openpdksdocs} for enabling a fully open-source design and
verification toolchain. We gratefully acknowledge
researchers from Shanghai Artificial Intelligence Laboratory and Fudan University for their helpful advice.

\bibliographystyle{plainnat}
\begingroup\setlength{\bibsep}{0pt}
\bibliography{references}

@misc{sky130docs,
  author = {{SkyWater PDK Authors}},
  title = {{SkyWater SKY130 PDK Documentation}},
  year = {n.d.},
  howpublished = {\url{https://skywater-pdk.readthedocs.io/en/main/}},
  note = {Accessed October 20, 2026}
}

@misc{skywatertechlef,
  author = {{SkyWater PDK Authors}},
  title = {{SKY130 High Density Standard Cell Technology LEF}},
  year = {n.d.},
  howpublished = {Public technology file},
  note = {\href{https://github.com/google/skywater-pdk-libs-sky130_fd_sc_hd/blob/main/tech/sky130_fd_sc_hd.tlef}{Official source}; DC-average current limits at a junction temperature of 90 degrees Celsius. Accessed November 12, 2026}
}

@misc{harbor,
  author = {{Harbor Contributors}},
  title = {{Harbor Documentation}},
  year = {n.d.},
  howpublished = {\url{https://docs.harborframework.com/}},
  note = {Accessed October 20, 2026}
}

@misc{dockerdocs,
  author = {{Docker, Inc.}},
  title = {{Docker Documentation: What Is Docker?}},
  year = {n.d.},
  howpublished = {\href{https://docs.docker.com/get-started/docker-overview/}{Official documentation}},
  note = {Accessed November 3, 2026}
}

@misc{gdsfactorydocs,
  author = {{gdsfactory community}},
  title = {{gdsfactory Documentation}},
  year = {n.d.},
  howpublished = {\href{https://gdsfactory.github.io/gdsfactory/}{Official documentation}},
  note = {Accessed November 3, 2026}
}

@misc{klayoutdocs,
  author = {{KLayout Project}},
  title = {{KLayout User Manual}},
  year = {n.d.},
  howpublished = {\href{https://www.klayout.de/doc/manual/index.html}{Official documentation}},
  note = {Accessed November 3, 2026}
}

@misc{magicdocs,
  author = {{Magic Development Team}},
  title = {{Magic VLSI Layout Tool}},
  year = {n.d.},
  howpublished = {\href{https://opencircuitdesign.com/magic/}{Official documentation}},
  note = {Accessed November 3, 2026}
}

@misc{ngspicedocs,
  author = {{ngspice Developers}},
  title = {{ngspice Documentation}},
  year = {n.d.},
  howpublished = {\href{https://ngspice.sourceforge.io/docs.html}{Official documentation}},
  note = {Accessed November 3, 2026}
}

@misc{analogdesignbench2026,
  author = {{Analog Design Bench Team}},
  title = {Long-Horizon Analog Design Bench: Benchmarking Agents on Hours-Long Analog and Mixed-Signal Circuit Design Tasks},
  year = {2026},
  howpublished = {arXiv:2609.33356v1},
  note = {\url{https://arxiv.org/abs/2609.33356v1}}
}

@inproceedings{jimenez2024swebench,
  author = {Jimenez, Carlos E. and Yang, John and Wettig, Alexander and Yao, Shunyu and Pei, Kexin and Press, Ofir and Narasimhan, Karthik},
  title = {{SWE-bench}: Can Language Models Resolve Real-World {GitHub} Issues?},
  booktitle = {International Conference on Learning Representations},
  year = {2024},
  note = {\href{https://proceedings.iclr.cc/paper_files/paper/2024/hash/edac78c3e300629acfe6cbe9ca88fb84-Abstract-Conference.html}{Conference paper}}
}

@inproceedings{liu2024agentbench,
  author = {Liu, Xiao and Yu, Hao and Zhang, Hanchen and Xu, Yifan and Lei, Xuanyu and Lai, Hanyu and Gu, Yu and Ding, Hangliang and Men, Kaiwen and Yang, Kejuan and Zhang, Shudan and Deng, Xiang and Zeng, Aohan and Du, Zhengxiao and Zhang, Chenhui and Shen, Sheng and Zhang, Tianjun and Su, Yu and Sun, Huan and Huang, Minlie and Dong, Yuxiao and Tang, Jie},
  title = {{AgentBench}: Evaluating {LLMs} as Agents},
  booktitle = {International Conference on Learning Representations},
  year = {2024},
  note = {\url{https://arxiv.org/abs/2308.03688}}
}

@inproceedings{zhou2024webarena,
  author = {Zhou, Shuyan and Xu, Frank F. and Zhu, Hao and Zhou, Xuhui and Lo, Robert and Sridhar, Abishek and Cheng, Xianyi and Ou, Tianyue and Bisk, Yonatan and Fried, Daniel and Alon, Uri and Neubig, Graham},
  title = {{WebArena}: A Realistic Web Environment for Building Autonomous Agents},
  booktitle = {International Conference on Learning Representations},
  year = {2024},
  note = {\url{https://arxiv.org/abs/2307.13854}}
}

@inproceedings{xie2024osworld,
  author = {Xie, Tianbao and Zhang, Danyang and Chen, Jixuan and Li, Xiaochuan and Zhao, Siheng and Cao, Ruisheng and Toh, Jing Hua and Cheng, Zhoujun and Shin, Dongchan and Lei, Fangyu and Liu, Yitao and Xu, Yiheng and Zhou, Shuyan and Savarese, Silvio and Xiong, Caiming and Zhong, Victor and Yu, Tao},
  title = {{OSWorld}: Benchmarking Multimodal Agents for Open-Ended Tasks in Real Computer Environments},
  booktitle = {Advances in Neural Information Processing Systems},
  volume = {37},
  year = {2024},
  note = {\url{https://arxiv.org/abs/2404.07972}}
}

@misc{merrill2026terminalbench,
  author = {Merrill, Mike A. and Shaw, Alexander G. and Carlini, Nicholas and others},
  title = {{Terminal-Bench}: Benchmarking Agents on Hard, Realistic Tasks in Command Line Interfaces},
  year = {2026},
  howpublished = {arXiv preprint arXiv:2601.11868},
  note = {\url{https://arxiv.org/abs/2601.11868}}
}

@misc{marten2026terminalbench4,
  author = {Marten, Ryan},
  title = {{Terminal-Bench 4.0}},
  year = {2026},
  howpublished = {Official benchmark release},
  note = {\url{https://www.tbench.ai/news/terminal-bench-4-0}, accessed November 3, 2026}
}

@inproceedings{liu2023verilogeval,
  author = {Liu, Mingjie and Pinckney, Nathaniel and Khailany, Brucek and Ren, Haoxing},
  title = {Invited Paper: {VerilogEval}: Evaluating Large Language Models for {Verilog} Code Generation},
  booktitle = {2023 IEEE/ACM International Conference on Computer Aided Design},
  year = {2023},
  doi = {10.1109/ICCAD57390.2023.10323812},
  note = {\url{https://arxiv.org/abs/2309.07544}}
}

@inproceedings{leibl2026loadbench,
  author = {Leibl, Markus and Azevedo, Filipe and Uhlich, Stefan and Bonetti, Andrea and Venkitaraman, Arun and Hsieh, Chia-Yu and Pham, Phuoc and Servadei, Lorenzo and Graeb, Helmut and Martins, Ricardo},
  title = {{LOADBench}: A Large Open Analog Dataset and Benchmark for Machine Learning in {IC} Design},
  booktitle = {Proceedings of the 2026 ACM/IEEE International Symposium on Machine Learning for CAD},
  year = {2026},
  doi = {10.1145/3831599.3840321},
  note = {\url{https://arxiv.org/abs/2610.21591v1}}
}

@inproceedings{xu2019magical,
  author = {Xu, Biying and Zhu, Keren and Liu, Mingjie and Lin, Yibo and Li, Shaolan and Tang, Xiyuan and Sun, Nan and Pan, David Z.},
  title = {{MAGICAL}: Toward Fully Automated Analog {IC} Layout Leveraging Human and Machine Intelligence},
  booktitle = {Proceedings of the 2019 IEEE/ACM International Conference on Computer-Aided Design},
  year = {2019},
  doi = {10.1109/ICCAD45719.2019.8942060},
  note = {Invited paper}
}

@article{dhar2021align,
  author = {Dhar, Tonmoy and Kunal, Kishor and Li, Yaguang and Madhusudan, Meghna and Poojary, Jitesh and Sharma, Arvind K. and Xu, Wenbin and Burns, Steven M. and Harjani, Ramesh and Hu, Jiang and Kirkpatrick, Desmond A. and Mukherjee, Parijat and Yaldiz, Soner and Sapatnekar, Sachin S.},
  title = {{ALIGN}: A System for Automating Analog Layout},
  journal = {IEEE Design \& Test},
  volume = {38},
  number = {2},
  pages = {8--18},
  year = {2021},
  doi = {10.1109/MDAT.2020.3042177},
  note = {Preprint: \url{https://arxiv.org/abs/2008.10682}}
}

@inproceedings{hammoud2024glayout,
  author = {Hammoud, Ali and Goyal, Chetanya and Pathen, Sakib and Dai, Arlene and Li, Anhang and Kielian, Gregory and Saligane, Mehdi},
  title = {Human Language to Analog Layout Using {GLayout} Layout Automation Framework},
  booktitle = {Proceedings of the 2024 ACM/IEEE International Symposium on Machine Learning for CAD},
  year = {2024},
  doi = {10.1145/3670474.3685971},
  note = {\url{https://openreview.net/pdf?id=4ZzkZkequO}}
}

@article{liu2024layoutcopilot,
  author = {Liu, Bingyang and Zhang, Haoyi and Gao, Xiaohan and Kong, Zichen and Tang, Xiyuan and Lin, Yibo and Wang, Runsheng and Huang, Ru},
  title = {{LayoutCopilot}: An {LLM}-powered Multi-agent Collaborative Framework for Interactive Analog Layout Design},
  journal = {IEEE Transactions on Computer-Aided Design of Integrated Circuits and Systems},
  volume = {44},
  number = {8},
  pages = {3126--3139},
  year = {2025},
  doi = {10.1109/TCAD.2025.3529805}
}

@misc{xu2026lamp,
  author = {Xu, Zecheng and Chen, Guanhua and Wu, Yumao and Zhu, Keren},
  title = {{LAMP}: {LLM}-Driven Analog Placement through Adaptive Tool Composition},
  year = {2026},
  howpublished = {arXiv preprint arXiv:2610.39882},
  note = {\url{https://arxiv.org/abs/2610.39882}}
}

@misc{chu2026codebehindlayout,
  author = {Chu, Minxing and Li, Anhang and Atilla, Martins and Qi, Ruichen and Saligane, Mehdi},
  title = {The Code Behind the Layout: {AI} Agents for Analog Layout Generators},
  year = {2026},
  howpublished = {arXiv preprint arXiv:2610.38426},
  note = {\url{https://arxiv.org/abs/2610.38426}}
}

@misc{saxena2026opera,
  author = {Saxena, Vishal and Mandal, Soumyajit},
  title = {{OPERA}: {OPamp} Design by an Expert-Taught Reasoning Agent, from Specifications to Layout},
  year = {2026},
  howpublished = {arXiv preprint arXiv:2610.39395},
  note = {\url{https://arxiv.org/abs/2610.39395}}
}

@misc{cadence2026virastack,
  author = {{Cadence Design Systems}},
  title = {{ViraStack AI Super Agent} for Custom and Analog {IC} Design},
  year = {2026},
  howpublished = {Product documentation},
  note = {\href{https://www.cadence.com/en_US/home/tools/custom-ic-analog-rf-design/virastack-ai-super-agent.html}{Official product page}, accessed November 3, 2026}
}

@misc{liu2026icpi,
  author = {Liu, Bingyang and Wei, Ziming and Gao, Xiaohan and Pan, David Z.},
  title = {Simulation-Aware In-Context Policy Improvement for {LLM}-Aided Analog Layout Refinement},
  year = {2026},
  howpublished = {arXiv:2608.13767},
  note = {\url{https://arxiv.org/abs/2608.13767}}
}

@misc{zhang2026panda,
  author = {Zhang, Haoyi and Fan, Weijian and Gao, Xiaohan and Liu, Bingyang and Wang, Runsheng and Lin, Yibo},
  title = {{PANDA}: An {LLM}-Enhanced Performance-Driven Analog Design Framework Bridging Design Intent and Layout Generation},
  year = {2026},
  howpublished = {arXiv:2606.15052},
  note = {\url{https://arxiv.org/abs/2606.15052}}
}

@misc{yoshioka2026vibetapeout,
  author = {Yoshioka, Kentaro},
  title = {The Era of Vibe Tapeout: Dark Age or Golden Age for Analog Circuit Design?},
  year = {2026},
  howpublished = {arXiv:2610.15240},
  note = {\url{https://arxiv.org/abs/2610.15240}}
}

@misc{tapeoutbench2026,
  author = {{Simra AI}},
  title = {{TapeoutBench} v0.1: Coding Agents on Thirteen Steps of a Chip Tapeout},
  year = {2026},
  howpublished = {Technical report},
  note = {\url{https://tapeoutbench.simratech.com/v0.1/}, accessed October 20, 2026}
}

@misc{chipagents2026buildchip,
  author = {Gunukula, Surya and Wang, Jerry Yicheng and Zhang, Kevin},
  title = {Introducing {Build-A-Chip-Bench}: From Workload Specification to {GDS}},
  year = {2026},
  howpublished = {ChipAgents Research Blog},
  note = {\url{https://chipagents.ai/research/build-a-chip-bench}, accessed November 3, 2026}
}

@misc{li2026internbootcamp,
  author = {Peiji Li and Jiasheng Ye and Yongkang Chen and Yichuan Ma and Zijie Yu and Kedi Chen and Xiaozhe Li and Ganqu Cui and Haozhan Li and Jiacheng Chen and Chengqi Lyu and Wenwei Zhang and Linyang Li and Qipeng Guo and Dahua Lin and Bowen Zhou and Kai Chen},
  title = {{InternBootcamp Technical Report: Boosting LLM Reasoning with Verifiable Task Scaling}},
  year = {2026},
  howpublished = {arXiv:2508.08636v2},
  note = {\url{https://arxiv.org/abs/2508.08636v2}}
}

@misc{li2026internagentharness,
  author = {Xiaozhe Li and Yongkang Chen and Shujian Deng and Peiji Li and Yichuan Ma and Huaxi Huang and Qiye Cai and Tianyi Lyu and Le Ma and Linyang Li and Qipeng Guo and Dahua Lin and Kai Chen},
  title = {{InternAgentHarness: A Scalable Synthetic Environment for Enhancing LLM Agentic Abilities}},
  year = {2026},
  howpublished = {Preprints.org},
  note = {\url{https://www.preprints.org/manuscript/202609.0958}}
}

@misc{codexcli,
  author = {{OpenAI}},
  title = {{Codex CLI}},
  year = {n.d.},
  howpublished = {\url{https://github.com/openai/codex}},
  note = {Accessed November 12, 2026}
}

@misc{opencode,
  author = {{Anomaly}},
  title = {{OpenCode}},
  year = {n.d.},
  howpublished = {\url{https://github.com/anomalyco/opencode}},
  note = {Accessed November 12, 2026}
}

@misc{claudecode,
  author = {{Anthropic}},
  title = {{Claude Code}},
  year = {n.d.},
  howpublished = {\url{https://github.com/anthropics/claude-code}},
  note = {Accessed November 12, 2026}
}

@misc{kimicode,
  author = {{Moonshot AI}},
  title = {{Kimi Code CLI}},
  year = {n.d.},
  howpublished = {\url{https://github.com/MoonshotAI/kimi-code}},
  note = {Accessed November 12, 2026}
}

@misc{qwencode,
  author = {{Qwen Team}},
  title = {{Qwen Code}},
  year = {n.d.},
  howpublished = {\url{https://github.com/QwenLM/qwen-code}},
  note = {Accessed November 12, 2026}
}

@misc{deepseek-harness2026,
  title = {{DeepSeek Harness: Everything is a Plugin}},
  author = {{DeepSeek-AI}},
  year = {2026},
  publisher = {GitHub},
  howpublished = {\url{https://github.com/deepseek-ai/deepseek-harness}}
}

@misc{openai2026gpt61sol,
  author = {{OpenAI}},
  title = {{Addendum to GPT-6 Astra System Card: GPT-6.1 Sol}},
  year = {2026},
  howpublished = {System card addendum},
  note = {September 29. \url{https://deploymentsafety.openai.com/gpt-6-1-sol}}
}

@misc{openai2026gpt6astra,
  author = {{OpenAI}},
  title = {{GPT-6 Astra System Card}},
  year = {2026},
  howpublished = {System card},
  note = {September 3. \url{https://deploymentsafety.openai.com/gpt-6-astra}}
}

@misc{deepseekai2026deepseekv41flash,
  author = {{DeepSeek-AI}},
  title = {{DeepSeek-V4.1-Flash: Pushing the Limits of KV Cache Compression}},
  year = {2026},
  howpublished = {Technical report},
  note = {\url{https://huggingface.co/deepseek-ai/DeepSeek-V4.1-Flash/blob/main/DeepSeek_V41_Tech_Report.pdf}}
}

@misc{deepseekai2026deepseekv4,
  author = {{DeepSeek-AI}},
  title = {{DeepSeek-V4: Towards Highly Efficient Million-Token Context Intelligence}},
  year = {2026},
  howpublished = {arXiv:2606.19348},
  note = {\url{https://arxiv.org/abs/2606.19348}}
}

@misc{glm5team2026glm5vibecodingagentic,
  title = {{GLM-5}: from Vibe Coding to Agentic Engineering},
  author = {{GLM-5-Team}},
  year = {2026},
  eprint = {2602.15763},
  archivePrefix = {arXiv},
  primaryClass = {cs.LG},
  url = {https://arxiv.org/abs/2602.15763}
}

@misc{moonshot2026kimik26,
  title = {{Kimi K2.6}: Advancing Open-Source Coding},
  author = {{Moonshot AI}},
  year = {2026},
  month = apr,
  howpublished = {Official technical blog},
  url = {https://www.kimi.com/en/blog/kimi-k2-6}
}

@misc{kimiteam2026kimik3,
  title = {{Kimi K3}: Open Frontier Intelligence},
  author = {{Kimi Team}},
  year = {2026},
  eprint = {2607.24653},
  archivePrefix = {arXiv},
  primaryClass = {cs.CL},
  url = {https://arxiv.org/abs/2607.24653}
}

@techreport{qwen2026design,
  title = {On the Design of {Qwen3.8-Next} Architecture: Evaluation, Efficiency, and Training Stability},
  author = {{Qwen Team}},
  institution = {Alibaba Group},
  year = {2026},
  month = aug,
  eprint = {2608.30320},
  archivePrefix = {arXiv},
  url = {https://arxiv.org/abs/2608.30320}
}

@misc{li2026agentorchestration,
  author = {Linyang Li},
  title = {{Agent-Orchestration in Autonomous Chip Design}},
  year = {2026},
  howpublished = {arXiv:2608.14035},
  note = {\url{https://arxiv.org/abs/2608.14035}}
}

@inproceedings{zheng2024sglang,
  author = {Lianmin Zheng and Liangsheng Yin and Zhiqiang Xie and Chuyue Sun and Jeff Huang and Cody Hao Yu and Shiyi Cao and Christos Kozyrakis and Ion Stoica and Joseph E. Gonzalez and Clark Barrett and Ying Sheng},
  title = {{SGLang}: Efficient Execution of Structured Language Model Programs},
  booktitle = {Advances in Neural Information Processing Systems},
  volume = {37},
  year = {2024},
  note = {\url{https://proceedings.neurips.cc/paper_files/paper/2024/hash/724be4472168f31ba1c9ac630f15dec8-Abstract-Conference.html}}
}

@misc{netgendocs,
  author = {{Netgen Developers}},
  title = {{Netgen 1.5: Netlist Comparison (LVS) and Format Manipulation}},
  year = {n.d.},
  howpublished = {\url{https://opencircuitdesign.com/netgen/welcome.html}},
  note = {Accessed November 12, 2026}
}

@misc{openpdksdocs,
  author = {{open\_pdks Contributors}},
  title = {{open\_pdks: Open Process Design Kits}},
  year = {n.d.},
  howpublished = {\url{https://github.com/fossi-foundation/open-pdks}},
  note = {Accessed November 12, 2026}
}

@misc{anthropic2026claudeopus55,
  author = {{Anthropic}},
  title = {{Claude Opus 5.5 System Card}},
  year = {2026},
  howpublished = {System card},
  note = {September. \url{https://www.anthropic.com/claude-opus-5-5-system-card}}
}

@misc{interns1team2025interns1,
  author = {{Intern-S1 Team}},
  title = {{Intern-S1: A Scientific Multimodal Foundation Model}},
  year = {2025},
  eprint = {2508.15763},
  archivePrefix = {arXiv},
  primaryClass = {cs.LG},
  note = {\url{https://arxiv.org/abs/2508.15763}}
}

@techreport{johnson2015latchup,
  author = {Marty Johnson and Roger Cline and Scott Ward and Joe Schichl},
  title = {{Latch-Up}},
  institution = {Texas Instruments},
  number = {SCAA124},
  year = {2015},
  month = apr,
  note = {\url{https://www.ti.com/lit/wp/scaa124/scaa124.pdf}}
}

@inproceedings{jerke2014mission,
  author = {Goeran Jerke and Andrew B. Kahng},
  title = {{Mission Profile Aware IC Design---A Case Study}},
  booktitle = {Proceedings of the Design, Automation \& Test in Europe Conference \& Exhibition},
  pages = {1--6},
  year = {2014},
  doi = {10.7873/DATE.2014.077},
  note = {\url{https://vlsicad.ucsd.edu/Publications/Conferences/312/c312.pdf}}
}

@misc{li2026tlgrpo,
  author = {Peiji Li and Linyang Li and Handa Sun and Wenjin Mai and Yongkang Chen and Xiaozhe Li and Yue Shen and Yichuan Ma and Yiliu Sun and Jiaxi Cao and Zhishu He and Bo Wang and Xiaoqing Zheng and Zhaori Bi and Xipeng Qiu and Qipeng Guo and Kai Chen and Dahua Lin},
  title = {{TL-GRPO}: Turn-Level {RL} for Reasoning-Guided Iterative Optimization},
  year = {2026},
  howpublished = {arXiv:2601.16480},
  note = {\url{https://arxiv.org/abs/2601.16480}}
}

@misc{kicadschematic9,
  author = {{KiCad Documentation Contributors}},
  title = {{KiCad} Schematic Editor Reference Manual, Version 9.0},
  year = {n.d.},
  howpublished = {Official documentation},
  url = {https://docs.kicad.org/9.0/en/eeschema/eeschema.html},
  note = {Simulator screenshot cropped for Figure 3. Creative Commons Attribution 3.0 or later (https://creativecommons.org/licenses/by/3.0/). Accessed November 12, 2026}
}
\endgroup

\clearpage
\appendix
\section{Contributions}
The listing of contributors is in alphabetical order based on their last names.
\par\vspace{16pt}
\begingroup
\renewcommand{\thefootnote}{\fnsymbol{footnote}}%
\makeatletter
\renewcommand{\subsection}{%
  \@startsection{subsection}{2}{\z@}{-20pt}{10pt}%
    {\normalsize\bfseries\raggedright}%
}
\makeatother

\subsection{Project Lead}
Linyang~Li

\subsection{Core Contributors}
Peiji~Li\footnote[1]{First authors. Equal contribution.},
Jianqiu~Chen\footnotemark[1], Wenjin~Mai

\subsection{Contributors}
Jie~Chen, Yongkang~Chen, Hongwei~Fang, Zhishu~He, Siqi~Li, Jianze~Liang,
Jialin~Lu, Zeyao~Lu, Yue~Shen, Xun~Wang, Jianxi~Xue

\par
\endgroup

\section{Limitations and Future Work}
\label{app:limitations}

\paragraph{Task and process coverage.}
The current release covers 40 fixed-netlist tasks using the SKY130 PDK.
It evaluates physical implementation with predefined devices and
specifications; topology selection and transistor sizing are outside the
task scope. Results characterize performance within the included circuit
families and scale range. Extending the suite to additional process
technologies and larger hierarchical circuits is an important direction
for future work.

\paragraph{Verification boundary.}
Acceptance is defined by the design rule decks, device models,
extraction settings, and electrical test cases. These checks provide
reproducible rule-based evidence under the stated conditions, but do not establish
foundry sign-off or guarantee silicon performance. The study includes no
silicon measurements, and performance under untested conditions remains
unverified.

\paragraph{Evaluation of layout quality.}
Passing DRC, LVS, and the prescribed post-layout simulations is necessary
for benchmark acceptance, but is far from sufficient to characterize
practical layout quality. We introduce geometric and engineering-quality metrics to narrow
this gap. These metrics were designed in consultation with analog layout
engineers. As an initial
approximation, these metrics may have incomplete coverage or assign low
scores to otherwise sound layouts. We will refine these metrics through
expert review and welcome community feedback and suggestions.

\clearpage
\section{Evaluation Details}
\label{app:evaluation}

\paragraph{Parameters and device area.}
The weights in Equation~\ref{eq:layout-score}, ordered as GDS, LVS, DRC,
area, quality, and simulation, are $(0.10,0.10,0.10,0.20,0.10,0.40)$;
$\gamma=2$. Engineers specify $q_t^{\max}\in\{3,4\}$ and
$\rho_t^\star\in[0.60,1.10]$; all $\rho_t^{\min}=0.10$
(Table~\ref{tab:task-parameters}). $A_{\mathrm{eff}}$ requires complete reference-device correspondence:
MOS contributions are $WL+A_{\mathrm S}+A_{\mathrm D}$; resistors and
capacitors use native extracted area parameters; BJT contributions union
collector/base/emitter recognition regions within each device. Physical-device
areas are summed without reapplying multiplicity or finger count.
Routing, gate extensions outside diffusion, substrate, and PCell envelopes
are excluded. Different devices may overlap, allowing $\rho_t>1$.

\paragraph{Engineering quality.}
Task annotations and LVS correspondence support three categories: matching
(placement relations, local environments, orientation, and array ends),
critical routing (detours relative to a spanning-tree lower bound and
differential balance), and power delivery (rail geometry, via redundancy,
and, where warranted by currents, resistive-grid estimates of wire/via stress
and voltage drop). These are layout proxies, not reliability sign-off.
For applicable categories $j\in\mathcal J_t$, raw scores
$r_{tj}\in[0,1]$ yield
\begin{equation}
Q_t=(1-p_t)\,
\frac{\sum_{j\in\mathcal J_t}
\alpha_j\min\!\left\{1,\frac{r_{tj}}{\tau_j}\right\}}
{\sum_{j\in\mathcal J_t}\alpha_j}.
\label{eq:engineering-quality}
\end{equation}
In matching/routing/power order, weights $\alpha_j$ are $(0.40,0.45,0.15)$
and saturation targets $\tau_j$ are $(0.80,0.80,0.90)$.
The geometric coupling penalty $p_t\in[0,1]$ concerns annotated
aggressor--victim nets. Inapplicable categories are omitted; missing
applicable measurements retain their weight with value zero.
Set $Q_t=0$ if $\mathcal J_t$ is empty, and $p_t=1$ if required coupling
evidence is unavailable.

\paragraph{Morphology and backend conventions.}
The screen rejects the joint presence of $\ge5$ long parallel signal nets,
$\ge5$ perpendicular drops, and a device row with length/thickness $\ge4$ containing $\ge70\%$
of matched devices. All conditions must hold;
unknown morphology withholds eligibility.

\paragraph{Electrical credit.}
Electrical credit is based on post-layout simulation with extracted parasitics.
For the fixed electrical verification manifest $\mathcal C_t$,
$F_t=I_{\mathrm{PEX}}|\mathcal C_t|^{-1}\sum_{c\in\mathcal C_t}I_c$,
where $I_{\mathrm{PEX}}$ denotes successful extraction; $I_c=1$ only for a
completed item satisfying all required checks with finite measurements.
Unfinished items remain in the denominator and verified passes survive
timeouts. Missing evidence yields a lower-bound score, not a presumed pass.

\begin{table}[!htbp]
  \centering
  \caption{Task-specific geometry parameters for all 40 tasks; $\rho_t^{\min}=0.10$ for all tasks.}
  \label{tab:task-parameters}
  \footnotesize
  \setlength{\tabcolsep}{4pt}
  \begin{tabular}{rrr@{\hspace{5pt}{\color{black!30}\vrule width 0.35pt}\hspace{5pt}}rrr@{\hspace{5pt}{\color{black!30}\vrule width 0.35pt}\hspace{5pt}}rrr@{\hspace{5pt}{\color{black!30}\vrule width 0.35pt}\hspace{5pt}}rrr@{\hspace{5pt}{\color{black!30}\vrule width 0.35pt}\hspace{5pt}}rrr}
    \toprule
    Task & $q_t^{\max}$ & $\rho_t^\star$ & Task & $q_t^{\max}$ & $\rho_t^\star$ & Task & $q_t^{\max}$ & $\rho_t^\star$ & Task & $q_t^{\max}$ & $\rho_t^\star$ & Task & $q_t^{\max}$ & $\rho_t^\star$ \\
    \midrule
    1 & 3 & 1.05 & 9 & 4 & 0.60 & 17 & 4 & 0.70 & 25 & 4 & 0.80 & 33 & 3 & 0.80 \\
    2 & 4 & 1.10 & 10 & 3 & 0.90 & 18 & 4 & 0.95 & 26 & 3 & 0.85 & 34 & 3 & 0.95 \\
    3 & 4 & 1.05 & 11 & 3 & 0.90 & 19 & 3 & 0.95 & 27 & 4 & 0.60 & 35 & 3 & 0.95 \\
    4 & 4 & 0.60 & 12 & 3 & 0.95 & 20 & 4 & 0.65 & 28 & 4 & 0.80 & 36 & 3 & 0.95 \\
    5 & 3 & 0.90 & 13 & 4 & 0.65 & 21 & 3 & 0.95 & 29 & 3 & 0.95 & 37 & 4 & 0.90 \\
    6 & 4 & 0.90 & 14 & 4 & 0.90 & 22 & 3 & 1.10 & 30 & 4 & 0.90 & 38 & 3 & 0.80 \\
    7 & 3 & 0.90 & 15 & 3 & 0.60 & 23 & 4 & 0.65 & 31 & 3 & 0.75 & 39 & 3 & 0.80 \\
    8 & 3 & 0.90 & 16 & 3 & 0.95 & 24 & 3 & 1.00 & 32 & 4 & 1.10 & 40 & 3 & 1.10 \\
    \bottomrule
  \end{tabular}
\end{table}

\clearpage
\section{Task Scale Analysis}
\label{app:task-scale}

\begin{figure}[!htbp]
  \centering
  \includegraphics[width=\linewidth]{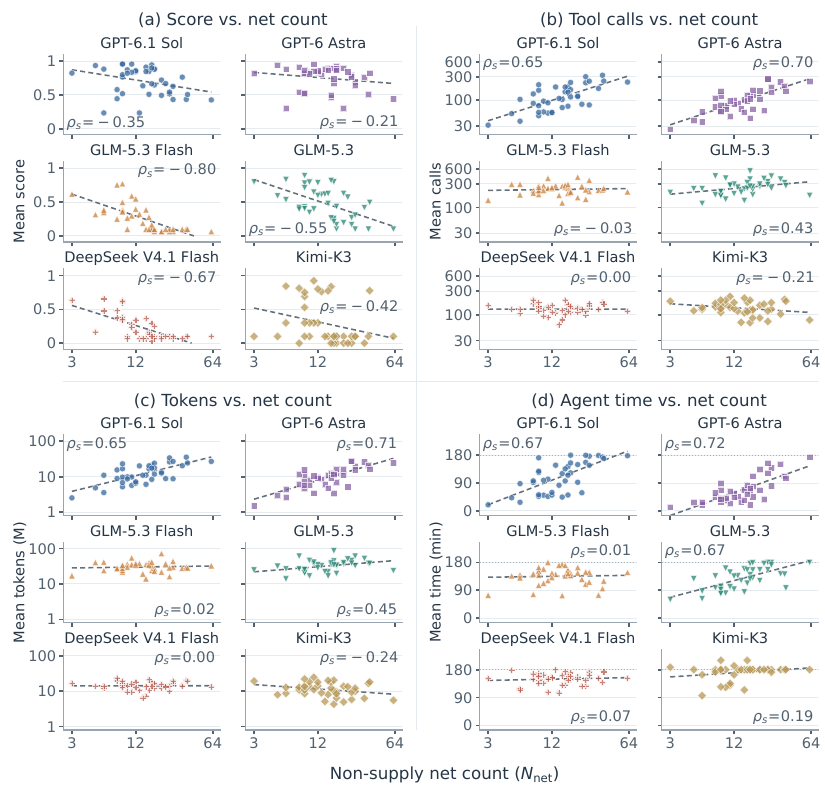}
  \caption{\textbf{Quality and resource use versus non-supply net count.}
  Groups show (a) task-level mean scores, (b) tool calls, (c) reported tokens,
  and (d) agent time.
  Configurations, scales, and line conventions match Figure~\ref{fig:task-scale};
  in-panel annotations report Spearman rank correlations.}
  \label{fig:task-scale-nets}
\end{figure}

We repeat the score and resource-use analyses using non-supply net count
(Figure~\ref{fig:task-scale-nets}). This counts distinct electrical nets
connected to devices after flattening the reference netlist, excluding
designated supply, ground, and substrate nets; each net is counted once,
regardless of how many terminals it connects. The trends remain similar
to the device-count analysis.

\clearpage
\section{Additional Case Studies}
\label{app:case-studies}
These examples separate compactness and circuit correctness using
individual final submissions evaluated under the same per-task parameters.

\subsection{Compactness at Comparable Device Content}
\label{app:case-compactness}
\begin{figure}[!htbp]
  \centering
  \includegraphics[width=\linewidth]{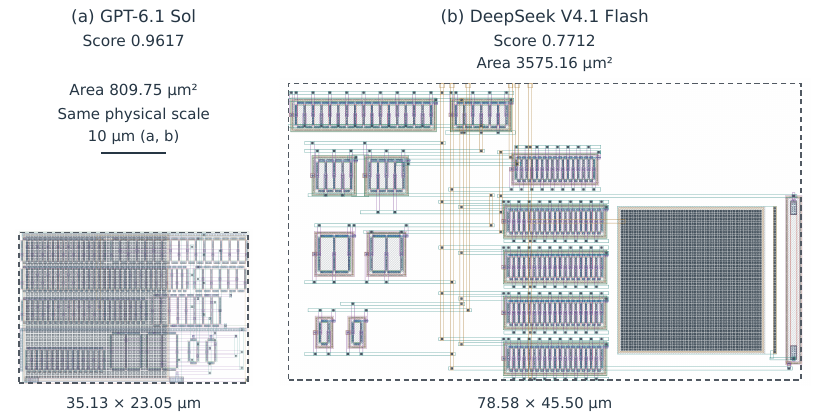}
  \caption{\textbf{Compactness control on the local two-stage OTA.}
  The two complete layouts share the same physical scale. Similar effective
  device areas and engineering-quality values accompany substantially
  different bounding areas, marked by dashed outlines. Both pass the
  prescribed physical and electrical checks including DRC, LVS and post-layout simulation.}
  \label{fig:case-compactness}
\end{figure}

On the local two-stage OTA (Task~40), GPT-6.1 Sol and DeepSeek V4.1 Flash both pass
the required physical and electrical checks
(Figure~\ref{fig:case-compactness}). Their effective device areas are
$910.01$ and $912.04\,\mu\mathrm m^2$, differing by only $0.22\%$.
Both earn 0.7 from submission, physical, and electrical verification,
and their engineering-quality contributions are similar, at 0.0617 and 0.0605.
Yet the bounding areas are $809.75$ and $3575.16\,\mu\mathrm m^2$.
The area-score contributions are consequently 0.2000 and 0.0108,
accounting for $99.4\%$ of the total-score difference
(0.9617 versus 0.7712). With nearly unchanged effective device content,
this pair makes the score's preference for compact physical organization
directly inspectable.

\clearpage
\subsection{Circuit Correctness beyond Visual Appearance}
\label{app:case-electrical}
\begin{figure}[!htbp]
  \centering
  \includegraphics[width=\linewidth]{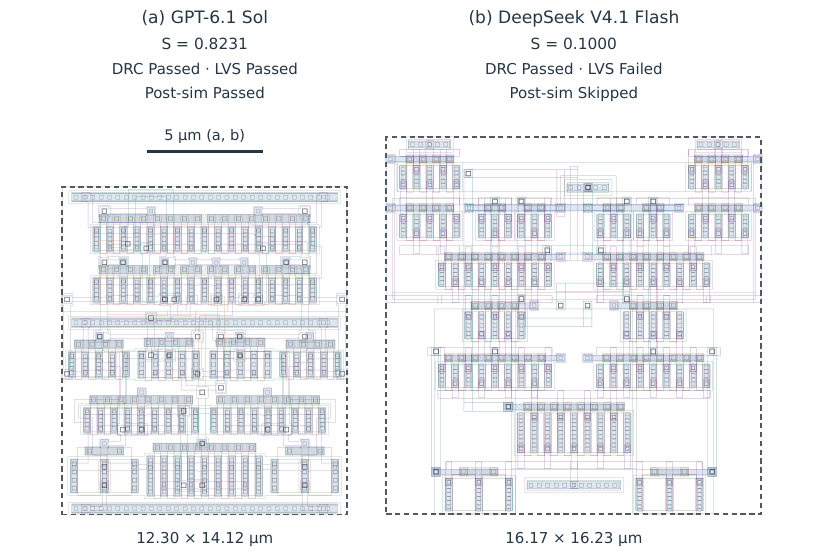}
  \caption{\textbf{Compact layouts with different circuit correctness.}
  The local-comparator submissions share a physical scale; dashed outlines
  mark their scoring bounds. Both pass DRC, but incorrect internal gate
  connections cause the DeepSeek layout to fail LVS. Its post-layout tests
  are skipped. The GPT-6.1 Sol submission passes all prescribed checks.}
  \label{fig:case-connectivity}
\end{figure}

Compact appearance and clean DRC do not establish circuit equivalence.
Figure~\ref{fig:case-connectivity} compares two submissions for the local
comparator (Task~38). Both pass DRC and form compact blocks, yet the DeepSeek V4.1 Flash
layout fails LVS after successful extraction. The extracted netlist reveals
incorrect gate connections in the cross-coupled NMOS pair: each gate is
connected to its local latch node rather than the opposite node required
by the reference. All twelve named top-level ports match, so this is an
internal connectivity error rather than a port-naming mismatch.

GPT-6.1 Sol passes LVS and all prescribed post-layout tests, scoring 0.8231:
0.3 for submission and physical verification, 0.0333 for area, 0.0898 for
engineering quality, and 0.4 for electrical verification. DeepSeek V4.1 Flash
receives only the 0.1 submission credit. Its failed LVS withholds DRC credit
and the remaining components; parasitic extraction and post-layout tests
are consequently skipped, rather than recorded as failed tests.
The contrast demonstrates why compact geometry must be checked against
the intended circuit before receiving layout-quality credit.

\end{document}